\pdfoutput=1
\documentclass[11pt]{article}

\usepackage[preprint]{acl}

\usepackage{times}
\usepackage{latexsym}
\usepackage{amssymb}
\usepackage{graphicx}
\usepackage{rotating}
\usepackage{comment}
\usepackage[T1]{fontenc}
\usepackage[utf8]{inputenc}

\usepackage{microtype}

\usepackage{inconsolata}

\usepackage{graphicx}
\usepackage{multirow}

\usepackage{amsmath}

\newcommand{\COL}{\texttt{[COL]}}
\newcommand{\CELL}{\texttt{[CELL]}}

\providecommand{\tabularnewline}{\\}

\title{Structural Deep Encoding for Table Question Answering}

\author{Raphaël Mouravieff \\
  Sorbonne Université, \\
  CNRS, ISIR, \\
  F-75005 Paris, France\\
  \texttt{raphael.mouravieff}\\\texttt{@isir.upmc.fr} \\ \\\And
  Benjamin Piwowarski \\
  Sorbonne Université, \\
  CNRS, ISIR, \\
  F-75005 Paris, France\\
  \texttt{benjamin.piwowarski}\\\texttt{@isir.upmc.fr} \\ \\\And
  Sylvain Lamprier \\
  LERIA, \\  Université d’Angers, \\ France  \\
  \texttt{sylvain.lamprier}\\\texttt{@univ-angers.fr} \\}

\begin{document}
\maketitle
\begin{abstract}

Although Transformers-based architectures excel at processing textual information, their naive adaptation for tabular data often involves flattening the table structure. This simplification can lead to the loss of essential inter-dependencies between rows, columns, and cells, while also posing scalability challenges for large tables. To address these issues, prior works have explored special tokens, structured embeddings, and sparse attention patterns. 
In this paper, we conduct a comprehensive analysis of tabular encoding techniques, which highlights the crucial role of attention sparsity in preserving structural information of tables. 
We also introduce a set of novel sparse attention mask designs for tabular data, that not only enhance computational efficiency but also preserve structural integrity, leading to better overall performance.

\end{abstract}

\section{Introduction}

Tabular data is a common data format \cite{cafarella2008webtables}, with many downstream tasks such as table question answering \cite{nan2022fetaqa} or table fact verification \cite{chen2019tabfact}. Tables present unique challenges to the research community due to their structured format in rows and columns, the heterogeneous nature of the data inside each cell, as well as the vast diversity of tables a model can encounter.
% Given that tables often contain both textual and numerical data, 
In the last few years, transformer models have become the dominant approach for modeling this format.

However, most works flatten a table into a sequence of tokens treating it as a linear text \cite{lu2024large}. This approach introduces critical limitations for the preservation of crucial structural features. %information. % , which is fundamental to accurately capture 
%relationships between data chunks. %rows, columns, and cells. 
Additionally, the quadratic complexity of the self-attention mechanism in transformers makes processing large tables computationally expensive~\cite{tay2022efficienttransformerssurvey}. Furthermore, it has been shown \cite{xie-etal-2022-unifiedskg} that a variety of these approaches suffer from over-fitting issues. %These limitations underscore the need for novel methods that can simultaneously preserve the structure of tables, reduce computational overhead, and improve generalization.

Several approaches have been proposed to address these challenges. One approach consists in introducing special tokens to explicitly mark rows and columns, as proposed in models like TAPEX and OmniTab \cite{liu2021tapex, jiang2022omnitab}. Other approaches propose to capture structural relationships  between table elements %, it is possible to 
by using structural embeddings \cite{herzig2020tapas,wang2021tuta}  or biasing attention~\cite{yang2022tableformer}.

Despite these advancements, limited research has explored how incorporating prior structural information affects the generalization performance of table processing models. We argue that sparse attention mechanisms, which have been used to manage table size in models like MATE~\cite{eisenschlos2021mate}, offer an untapped opportunity to leverage structural information. While MATE primarily focuses on managing table size, we propose extending sparse attention patterns to encode structural relationships within tables, enhancing both scalability and generalization capabilities. 
%We hypothesize that sparse attention mechanisms can significantly improve the generalization of tabular models by selectively focusing on key structural components, as well as 
%% By dynamically limiting attention to relevant regions of the table, sparse attention reduces the inclusion of irrelevant information and
%mitigating the computational overhead associated with global attention. 

In this paper, we systematically evaluate combinations of existing methods for preserving table structure and introduce new \textbf{sparse attention masks} specifically tailored to tabular data, as well as new modules designed to retain structural information.  Our contributions are threefold:
\begin{itemize}
    \item \textbf{Comprehensive Evaluation}: We systematically assess all existing table encoding techniques from the literature, as well as our newly introduced methods, across multiple dimensions of generalization.% This is the first work to provide a complete analysis of table encoding strategies.
    \item \textbf{Structural Encoding Guidelines}: Based on our findings, we provide practical recommendations for encoding table structure in transformer-based models. Specifically, we show that 1.)  Incorporating at least one form of absolute structural encoding enables models to learn table-specific rules, leading to improved generalization. 2.) Sparse attention masks, which selectively allow attention between specific table cells, significantly enhance generalization performance.
    \item \textbf{A Novel Sparse Attention Mask for Enhanced Efficiency}: We demonstrate that our proposed sparse attention masks, when integrated with optimized self-attention mechanisms, achieve significant computational speedups. These masks play a crucial role in scaling transformer-based models to efficiently handle large tables. % tabular datasets.
\end{itemize}

%we propose new sparse attention masks and modifications to classical encoding methods to further enhance structural preservation. Second, we systematically evaluate the different table encoding methodologies -- both from state-of-the-art models and our own, combining them when possible. We evaluate them across multiple dimensions of generalization—compositional, structural and robustness, providing a comprehensive analysis of how various structure preservation mechanisms impact model performance. Third, we offer critical insights into the architectural choices that lead to improved generalization, especially when handling large and complex tables. 
% Importantly, no prior study has explored the impact of integrating structural information into transformers on the generalization capabilities of tabular models, making our work a novel contribution to this growing field.

\section{Related Work}

\begin{figure*}[tp]
  \centering
  \includegraphics[width=0.8\linewidth]{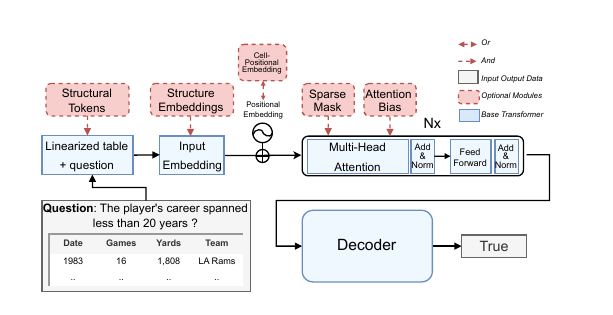}
  \caption{Overview of the encoding pipeline (blue) along with the different steps where table-specific information can be injected (red)}%the different approaches for Table QA and their limits, % (a-c), along with our proposition (d)}
  \label{fig:model}
\end{figure*}

\subsection{Table  models limitations}
Current table processing models exhibit several key limitations that make them challenging to deploy in real-world applications.\\
\textbf{Generalization Issue:} First, table models do not generalize well. As noted in~\cite{xie-etal-2022-unifiedskg}, table question-answering are easily perturbed by simple structural modifications, such as row or column permutations. In response to these challenges, new datasets focusing on robustness have emerged~\cite{zhou-etal-2024-freb,zhao-etal-2023-robut}. These datasets introduce other perturbations to table structures, further exposing the severe generalization limitations of state-of-the-art models, but do not explain the causes.
% While these datasets emphasize the problem, they do not attempt to explain the underlying causes of the generalization issues as we do in this paper.
%In this paper, we conduct an in-depth analysis and argue that the well-known generalization problem originates from the structural encoding choice used in table-based models. 
\\
\textbf{Table Size Issue}: The quadratic complexity of the self-attention mechanism in transformers poses a significant limitation when dealing with large inputs~\cite{tay2022efficienttransformerssurvey, ye2023large}. Models struggle to process large tables efficiently and tables must be truncated, leading to sub-optimal performance. \cite{ye2023large} further emphasizes these challenges, showcasing the difficulties transformers face when handling extensive table  data. 
Research from \cite{patnaik2024cabinet} proposes addressing this problem by weighting relevant parts of the table related to the answer, but this solution requires a complex pipeline with two interacting models. Another solution, sparse attention, as proposed in MATE~\cite{eisenschlos2021mate}, allows handling of larger tables. Their work however did not explore many sparsity patterns and their interaction with other factors.\\
%In this study, we examine different sparse attention patterns and demonstrate that certain patterns achieve similar generalization performance  while offering significantly improved sparsity.\\
\textbf{LLM Problems:} The above observations can be extended to Large Language Models (LLMs). \cite{sui2024table} demonstrates that even billion-parameter LLMs struggle with simple tasks, such as counting rows in large tables.
Additionally, findings from \cite{liu-etal-2024-rethinking} show that structural variations in tables containing the same content can lead to a significant drop in model performance, particularly in symbolic reasoning tasks.
Our approach can assist recent advances in LLMs, which, although typically based on decoder-only architectures, are beginning to integrate a table-specific encoder ~\cite{zha2023tablegpt}.

\subsection{Leveraging table structure}
Numerous methods have been developed to preserve the inherent structure of tables. The predominant approach in the literature involves linearizing and concatenating the table with the query, while attempting to retain its structural information \cite{dong2022table}.\\
\textbf{Input Token Structure:} A widely used method introduces special tokens to signal the table structure \cite{liu2021tapex,jiang2022omnitab,lin2020bridging}. The idea behind this approach is to preserve the table’s structural information by adding special tokens to delimit columns, rows, and cells. These tokens allow the model to infer which cells correspond to specific rows or columns. 
% There are many variations of this token-based encoding method, and in this paper, we propose a novel approach that we systematically compare to existing methods in the literature.\\
\\
\textbf{Structured Embeddings:} Another approach involves incorporating specific segment embeddings to help models access and manipulate table structure information. \cite{shi2022generation,herzig2020tapas,wang2021tuta} propose to use column and row-specific embeddings to encode the position of each token within the table. \cite{wang2021tuta} extends this approach with tree-based embeddings to capture hierarchical relationships within the table’s structure, including multiple headers. 
%However, column and row embeddings are limited by the maximum table size that models have encountered during training. 
\\
\textbf{Bias Attention:} \cite{zayats2021representations,yang2022tableformer} propose biasing the attention mechanism to incorporate prior knowledge about table structure directly into the model’s attention layers. %How this bias combines with other mechanisms has not been studied though.
% This bias has been studied in depth and shows promise for improving model performance on structured data.
\\
\textbf{Sparse Attention:} The use of attention masks, that consider the structure of tables, has been proposed in \cite{eisenschlos2021mate}. Masking attention between tokens of different columns or rows allows to integrate table structure by architecture design. This was however  introduced in \cite{eisenschlos2021mate} for computational efficiency purposes only. %, without any consideration for generalization  abilities such process can benefit. %, rather than in token  
%To our knowledge, no paper has explored sparse attention as a standalone method to capture table structure through row and column permutation invariance. The only related work, proposed in \cite{eisenschlos2021mate}, studies sparse attention purely for computational efficiency.

In this paper, we break down all the methods discussed in this section into individual components and systematically assess their interactions. 
% combination of each. We also examine their interactions to determine which elements are the most important for integrating into transformer models to ensure robust generalization.

\subsection{Generalization experiments}

The generalization capabilities of transformers have been a topic of sustained interest within the research community \cite{hupkes2023taxonomy}. We list below the different challenges.\\
\textbf{Structural Generalization} examines how well a model adapts when changing its input structure. Previous works, such as \cite{saparina2024improving}, have focused on how small variations affect performance. However, few studies have extended this to larger tables as we do in this work.\\
%. We propose testing our models on tables larger than those used for training. \\
% new tables with architectures never seen during training to explore how they generalize to different Out Of Domain table sizes.\\
\textbf{Compositional Generalization} refers to the ability to recombine known elements to handle new inputs. While there has been substantial work in logical problem domains~\cite{dziri2024faith, zhang2022unveiling}, how to handle  data tables remains under-explored. To the extent of our knowledge, the only work on tables regarding compositional generalization is that of
\cite{rai2023improving}, who introduces %novel techniques aimed at enhancing compositional generalization by proposing
improved tokenization strategies for tables. % data.
% Our study is the first to investigate this type of generalization specifically in SQL code execution. 
\\
%In this work, we propose testing our models on a subset of queries to assess their compositional generalization abilities.\\
%\textbf{Lingual and Cross-Domain Generalization:} This type of generalization assesses how models perform when the language or domain of the input data shifts. \cite{gan2021exploring} addresses cross-domain generalization by proposing a variant of the Spider dataset. Building on these ideas, we manipulate the learned vocabulary to evaluate how transformer models handle  domain-specific variations in tabular data.\\
\textbf{Robustness of Generalization:} Robustness, particularly in response to adversarial perturbations, is a critical area of study in table  question-answering. \cite{zhao-etal-2023-robut, chang2023dr} have developed datasets to test specific robustness properties. In this work, we extend these efforts by systematically analyzing model sensitivity to cell repetition.
%of table-processing transformers' robustness across perturbation scenarios.

% Our work builds on previous research on the generalization analysis of tabular models but distinguishes itself by providing explanations and proposing solutions to address these generalization challenges.

% --------

\section{Models and Structural Encoding Components} %Methodology}

%\subsection{Overview of the Approach}

In this section, %we present a structured experimental plan designed to evaluate a comprehensive set of methods developed in the literature for preserving table structure in tabular models. %The goal is to systematically assess these methods and identify the most effective ones for enhancing generalization.
%We begin by
we first present %presenting
our backbone models before  %decomposing approaches into
specifying various independent methods to encode data tables (Figure \ref{fig:model}). %We adapt the notations used in the different papers to ensure their homogeneity.

%, each component can be independently added or removed from the base model, allowing for flexible experimentation. This modular decomposition enables a thorough evaluation of different combinations, providing valuable insights into how these techniques interact and contribute to the model’s ability to generalize across diverse table structures and content.
%Next, %we introduce our proposed methods, which include novel strategies for handling special tokens, refined embedding techniques, and a set of custom-designed sparse attention masks. These innovations are aimed at improving the model’s capacity to represent and retain the structural information of tables.
 %we describe the datasets specifically created for this analysis. These datasets have been carefully curated to test the model’s ability to generalize by maintaining structural fidelity across various generalization scenarios.

%\subsection{Decomposition of Modules}

\subsection{Backbone} We adopt BART \cite{lewis2019bart} as the baseline model for our experiments. Following \cite{liu2021tapex}, \cite{jiang2022omnitab} who proved that BART can obtain very strong results on table benchmarks.
%We adopt BART \cite{lewis2019bart} as the baseline model for our experiments. Following \cite{liu2021tapex}, \cite{jiang2022omnitab} who proved that BART can obtain very strong results on table benchmarks.
% While BART was not originally designed for tabular data, we explore several adaptations to equip it with table-specific structural awareness. To this end, 
We follow standard table linearization strategies, %, employed by every method from the litterature,
by concatenating rows sequentially, with the query pre-pended. To the best of our knowledge, this remains the only linearization approach proposed in the literature. 
Besides the methods described below, we modify BART using a Segment Embedding \cite{devlin2018bert} that distinguishes the question from the rest of the sequence (i.e., the table). 
For an illustrative example, refer to Figure \ref{fig:specialtokens} (in appendix).  
All models share this common structure
% These two elements—BART’s core architecture and the Segment Embeddings—serve as the backbone of our experiments,
ensuring a fair and consistent comparison across models.

\subsection{Special Tokens (T)}
\label{sec:TokStruct}
Special tokens can be introduced in the linearization to encode structural information in the input sequence. We experiment with three types: T0 (no tokens), T1 (Row Indexed - Cell Tokens), and a novel variant % proposition
T2 (Row - Column - Cell Tokens). 
T1, used in models like TAPEX \cite{liu2021tapex}, marks each start of row with \texttt{[ROW  n]}, where $n$ is the row number, and each new cell by %as well as
a cell separator \CELL{}. 
T2 is proposed to explore the need for absolute markers of row numbers, which can hinder generalization abilities regarding permutation invariance. It also introduces  %replaces it by
a column special token \COL{}, which can be leveraged by structural masks described below. %also introduces an additional  additional \ROW, \COL, and \CELL{} tokens to mark rows, columns, and cells, providing more structural information and maintaining row permutation invariance. 
%

%Structural tokens are special tokens inserted in the input sequence to encode structural information. In our experiments, we consider three types: T0 (no structural token inserted), T1 (Row Indexed - Cell Tokens) and T2 (ROW - Column - Cell Tokens). For an illustrative example, refer to Appendix Figure \ref{fig:specialtokens}.  
%T1 is the structural token scheme introduced in mutiple articles such as TAPEX \cite{liu2021tapex}, OMNITAB \cite{jiang2022omnitab}. It includes a different special row token for each row in the table, in addition to a special generic token that is leveraged as a cell separator.
%To provide more structural information and ensure row permutation invariance, we also explore the T2 scheme. This approach introduces a generic \ROW{} token at the beginning of each row, a generic \COL{} token for each column, and a generic \CELL{} token at the start of each cell. We adopt traditional linearization strategies from the literature, which sequentially concatenate rows and append the query at the start of the flattened table.

\subsection{Structural Embeddings (E)}

An alternative to using special tokens is to add structural Embeddings to different segments of the input. In this paper, we compare E0 (No Structural Embeddings) with E1 (Row Column Embeddings), which corresponds to the Structural Embeddings introduced in TAPAS~
\cite{herzig2020tapas}. 
In the latter, each token embedding $X_i$ is augmented with structural ones, such as:
$\tilde{X}i = X_i + E_{\text{row}}(r_i) + E_{\text{col}}(c_i)$, where $E_{\text{row}}$ and $E_{\text{col}}$ are learnable embeddings specific to the row and column indices, allowing the model to differentiate between cells based on their position. \\ %al context.\\

\subsection{Cell/Table Positional Embedding  (PE)}
We also explore two distinct approaches for positional embedding: \\
\textbf{Table Positional Embedding (TPE):} the standard approach used in transformers like BART \cite{lewis2019bart}, applying a global sequence of positional embeddings across the entire input sequence. \\
\textbf{Cell Positional Embedding (CPE):} is a table-specific method where the positional embeddings index resets after each cell~\cite{eisenschlos2021mate, yang2022tableformer}\\

Note that throughout this paper, we consistently distinguish between \textbf{absolute methods}, which assign fixed positional values to tokens based on their location in the input (TPE, E1, T1) and \textbf{relative methods}, which remain independent of row and column order (the rest).

\subsection{Bias Attention (B)}

%Attention 
We use a set of learnable biases $B$,  added to classical transformer's attention scores (before the softmax operation), to  
%A set of learnable bias terms $B$ is used to add a bias to the attention, allowing the model to 
incorporate %additional 
relational information from the table in contextual encoding. %This is done  %In that way, attention score of a token $j$ to token $i$ is defined as :  
%$\tilde{a_{j\rightarrow i}} = \frac{q(x_i) \cdot k(x_j)}{\sqrt{d_k}} + b_{i,j}$, with classical query $q$ and key $k$ values where . 
We use the code from TableFormer \cite{yang2022tableformer} for the bias creation, where the attention bias $B_{i,j}$ for the attention of $i$ onto $j$  depends on the relationship between tokens $i$ and $j$ (i.e. cell to column header, etc.). More details about attention biases are given in appendix  %and the full list of biases, refer to 
(section \ref{appendix:tableformer_bias}). We note B1 for the presence of Bias and B0 the absence.  \\

\subsection{Sparse Masking in Attention (M)}
%\textbf{Sparse Masking in Attention}:

We also experiment the use of sparse attention masks to restrict transformer's attention between components of the table, depending on its  structure. This is done by adding a mask $M$ to all attention scores, before the softmax function, with $M_{i,j}=-\infty$ if the attention between $i$ and $j$ is  masked. %Attention of each component from the input is defined as:  % items to specific components modifies the attention mechanism by applying sparse attention masks.
%$\tilde{A} = \frac{QK^T}{\sqrt{d_k}} + M$ where $M_{ij}=-\infty$ when the attention between $i$ and $j$ is masked.\\
%\subsection{Proposed techniques to preserve structure}
We propose six different sparse masks, ranging from M1 (the least sparse, corresponding to MATE) to M6 (the most sparse), with M0 representing the case where no sparse attention is applied. Some of these sparse masks necessitate the structural tokens T2  (section \ref{sec:TokStruct}). These masks are marked with \textcolor{red}{*} to differentiate them from the others. Full details of our mask schemes can be found in Table~\ref{tab:attention}, and a more visual example is provided in appendix (Figure \ref{appendix:sparsemask}).

\begin{table}[ht]
\centering
\resizebox{0.48\textwidth}{!}{
\begin{tabular}{|l|c|c|c|c|c|c|}
\hline
\textbf{Attention} & M1 & M2 & M3 & M4\textcolor{red}{*} & M5\textcolor{red}{*} & M6\textcolor{red}{*} \\
\hline
$q_{i} \leftrightarrow  w_{r,c,k}$ & X  &  X & X  & X & X & X \\\hline
$q_{i} \leftrightarrow  q_{i'}$ & X & X & X& X & X & X \\
\hline
$w_{r,c,k} \leftrightarrow w_{r',c,k'}$ & X & X &  & X &  & \\
\hline
$w_{r,c,k} \leftrightarrow w_{r,c',k'}$ & X & & X & & X & \\
\hline
$w_{r} \leftrightarrow w_{r,c,k}$ & & & & X & X & X \\\hline
$w_{c} \leftrightarrow w_{r,c,k}$ & & & & X & X & X \\\hline
$w_{r,c} \leftrightarrow w_{r,c,k}$  & & & & X & X & X\\\hline
$w_T \leftrightarrow w_{r,c,k}$ & & & & X & X & X\\
\hline
\end{tabular}}
\caption{Summary of allowed attention ($\leftrightarrow$) patterns for different sparse masks. We denote $q_i$ a query token, $w_{r,c,k}$ a token at position $k$ in the cell located row $r$ / column $c$. Then, $w_{r,c,k} \leftrightarrow w_{r',c,k'}$ indicates that attention is \emph{not masked} between tokens within the same column across different rows.
Additionally, for T2, $w_c$, $w_r$, $w_{r,c}$ and $w_T$  denote the column $c$, row $r$, cell $(r,c)$ and table tokens respectively.}
\label{tab:attention}
\end{table}

\section{Experimental Setup}
\label{sec:experimental-setup}

\begin{comment}
%In addition to the real-world datasets introduced later, 
Synthetic data is employed to conduct an in-depth impact analysis of the different table encoding techniques. These synthetic datasets are meticulously designed to evaluate their generalization capabilities by examining their robustness in the presence of missing values, structural alterations, compositional generalization of queries, and of resilience to correlations between cell contents prevalent in real-world datasets (hereafter referred to as mixability).
% Each of these aspects is defined in detail in the subsequent sections. % . Furthermore, we control the similarity parameter to adjust the transition matrix used in generating the tables, allowing us to simulate different levels of structural complexity and variability.

\end{comment}

In addition to the real-world datasets introduced later, we use synthetic data for a more fine-grained impact analysis of the encoding factors. These synthetic datasets are designed to assess the generalization capabilities of the models by evaluating their robustness to missing values, structural changes, compositional generalization of queries, and resilience to correlation between cell contents often found in real world datasets~(referred to as mixability in the following).

\subsection{Synthetic Data Generation}
%In order to assess the impact of factors of encoding in specific - well-controlled - settings, we propose a methodology of synthetic data generation. Next, different alterations of these initial data are proposed, to challenge generalization and robustness capabilities of the models. 

 %First, we generate 
 A %synthetic
 dataset $\mathcal{T}$ corresponds to a set of triplets $(T,Q,A)$,  where $T$ is a table, %each sample $T \in \mathcal{T}$ consists of a table $T = \left( (w_{r,c})_{c=1\ldots N^T_{col}} \right)_{r=1\ldots N^T_{row}}$ is a matrix of  values $w_{r,c}$,
 $q$ is an SQL query, and $A$ is the %a 
 target answer.\\
\textbf{Table Generation:} Each table $T$  is a matrix,   %$ ((w_{r,c,k})_{k})_{c=1\ldots N^T_{col}})_{r=1\ldots N^T_{row}}$, with 
 $N^T_{row}$ rows and $N^T_{col}$ columns, both uniformly sampled from $\mathcal{U}(6,8)$  %${6,7,8}$, i.e., $N^T_{row}, N^T_{col} \sim \mathcal{U}(6,8)$
 for each sample. 
Each cell $(r,c)$ in $T$ is a sequence of %1, 2 or 3 
tokens $w_{r,c,k}$, resulting from the tokenization of a  random 
integer, sampled from  %$w_{jk} \sim 
$\mathcal{U}(V)$ with $V = \{0, 1, \dots, 999\}$.
\\
\textbf{Query Generation:} SQL queries are based on 10 common patterns from SQUALL \cite{shi2020potential}, 
% focusing on reasoning tasks, 
focusing on simple selection tasks,
that are instantiated according to contents of data they are applied on. We use the Fuzzingbook library\footnote{\url{https://github.com/uds-se/fuzzingbook}} to generate a set of variations from each SQL query template. For details about the SQL templates used, please refer to the appendix (section \ref{appendix:sql_train}). It is worth noting that SQL is a natural choice for studying Table Question Answering due to the structured alignment between text and SQL queries \cite{wang2019learning}. %We also choose SQL since it is easier for a transformer to learn to answer them, and simple selection since these tasks are difficult enough to generalize, and are the basis of more complex processing.

\subsection{Disturbances from In-Domain Data}

\noindent \textbf{Structural Generalization}: To evaluate the model’s robustness to structural changes, we test its performance on tables of varying sizes, both larger and smaller than those seen during training. The model is tested on tables with dimensions outside the training range, i.e. $N^T_{row}, N^T_{col} \in \{4,5,9,10,11,12\}$. 
% This allows us to assess the model’s ability to generalize to unseen table structures, which is crucial for real-world applications with diverse table sizes.

\noindent \textbf{Consistency Robustness:} 
Token repetitions make the task harder since the model cannot rely on the token semantic.  In our experiments on consistency robustness, we select a random word $v_0$ from our vocabulary $v \in V$. For each cell, we replace its content with $v_0$ with a probability $R$. We use either $R=0.2$ or $R=0.4$  (with uniform probability).
%
%
%To assess robustness to repeated values, we introduce a repetition parameter $R$ that controls the probability of any element $t_{jk}$ being a repeated value. For each $t_{jk}$:\\
%$P(t_{jk} = \text{RepeatedWord}) = R, \quad P(t_{jk} = \text{RandomWord}) = 1 - R$ The model is trained with $R = 0$ (no repetition) and tested with $R = {0.2, 0.4, 0.6}$ to measure the effect of increasing repetition on generalization.\\
%\textbf{Vocabulary Shift} : The goal is to evaluate the model’s generalization when exposed to a new vocabulary distinct from the training set.\\
%Training Phase: The model is trained on tables where entries are integers sampled from $V_{\text{train}} = {0, 1, \dots, 999}$.\\
%Test Phase: The model is tested on datasets with altered vocabularies: \\
%$V_{\text{test1}} = {1000, 1001, \dots, 1999}$
%$V_{\text{test2}} = {1000, 1001, \dots, 9999}$
%This setup evaluates the model’s ability to generalize to a new but contiguous numeric vocabulary.\\

\noindent \textbf{Compositional Generalization:} 
During training, we use an ensemble of 10 SQL templates (see Section \ref{appendix:sql_train} in appendix). These templates include basic patterns such as \texttt{SELECT cx FROM table WHERE cy IN {...}} or \texttt{SELECT cx FROM table LIMIT k}.
To evaluate compositional generalization, we use queries that combine these known components in new ways, such as \texttt{SELECT cx FROM table WHERE cy IN 
    {...} LIMIT k}. This setup allows us to assess the model's ability to generalize to more complex queries by combining simpler ones. 
    
\noindent \textbf{Mixability Robustness:} \label{sec:mixability}
We use a parameter $S \in [0,1]$ to control how deterministic the table content generation is. With $S = 1$, a cell content if fully determined by the previous cells in the row, while for $S = 0$, we use a random generation (same table generation process as in the training set). We detail the generation procedure in appendix (Section \ref{sec:mixability-generation}).

\subsection{Real Datasets and Evaluation}

Finally, we use the following real-world datasets in our experiments. 
\textbf{WikiTableQuestions (WTQ)} \cite{pasupat2015compositional} is a challenging benchmark for table question answering, as it includes numerical reasoning questions, tables with missing values, and noisy columns containing a mix of text and numerical cells.
\textbf{WikiSQL (WSQL)} \cite{zhong2017seq2sql}  is another table-based question answering dataset, where we use the provided SQL supervision to train models, mapping tables and questions to execution results. This dataset is considered to be simpler than WikiTableQuestions, as it more closely follows structured SQL queries.
%\textbf{Synthetic Table Generalization}: In addition to the real-world datasets, we use synthetic data as presented in Section (cite). These data are designed to test the generalization capabilities of the models by evaluating their robustness, structural understanding, and compositional generalization. Furthermore, we control the similarity parameter to adjust the transition matrix used in generating the tables, allowing us to simulate different levels of structural complexity and variability.\\
%\textbf{Evaluation}: We evaluate our models using standard denotation accuracy (DA) for both real and synthetic datasets. This metrics checks if the predicted answer is equal to the target answer. 

\subsection{Training pipeline}
\label{section:training_pipeline}
We initialize our models using pre-trained BART weights \cite{lewis2019bart}. %We choose not to apply intermediate pre-training on tabular-related tasks, as proposed in the literature \cite{eisenschlos2020understanding}, since o
Our SQL execution procedure %already
serves as a common technique for intermediate pre-training on tabular data \cite{liu2021tapex}.
%without any intermediate pre-training as in \cite{eisenschlos2020understanding}. 
For our training procedure on artificial datasets, we run up to 200k steps with an early stopping patience of 15 to ensure convergence, as suggested by \cite{csordas2021devil}. We use a batch size of 8, a context length of 512, and a learning rate of $3 \times 10^{-5}$. For fine-tuning on real datasets, we adopt the same hyperparameters as described in \cite{liu2021tapex}.

%\subsection{Baselines}
%
%In line with the modules proposed in the literature presented in (cite), we selected the most representative models for each category. We compare our approach with the following baselines: TAPEX for their proposed input token structure (T1), TAPAS for their use of structural embeddings (REC), TableFormer for their attention bias mechanisms (NB, B), and MATE for their use of sparse masks (M1) to handle long input encoding.

\section{Experiments}

We evaluate our models using denotation accuracy (DA) on both real and synthetic datasets. Denotation accuracy measures correctness by comparing ground-truth and predicted outputs, considering a prediction correct if the sets of values match, i.e. irrespective of order.

\subsection{ANOVA Decomposition of Structural Factors in Tabular Encoding}

We use ANOVA to evaluate the impact of structural components (Table \ref{tab:exp:anova}).

% , a statistical framework for evaluating the impact of factors and their interactions on model accuracy. 
% We report statistical significance (p-values) in bold along with effect sizes ($\eta^2 \in [0,1]$), quantifying associations with accuracy. 
%The full table is available in Appendix \ref{appendix:anova}.

\noindent \textbf{Main Effects:}  Positional Embeddings (PE), and Tabular Structure Embeddings (E) significantly affect model performance (and have a strong interaction PE $\times$ E). Positional embeddings exhibit the strongest effect ($\eta^2\in [0.18, 0.27]$), where TPE consistently outperforms CPE. Tabular structure embeddings also improves accuracy (E1 $>$ E0,  $\eta^2\in [0.15, 0.30]$). In contrast, sparse tokens (T), Bias (B) and mask (M) -- when considered independently -- exhibit no significant impact on model performance. These results indicate that absolute table encoding methods, such as TPE and RCE, are critical for table-based question answering, as models struggle to generalize with purely relative encodings. We hypothesize that without CPE or TPE, the decoder struggles to locate relevant information.

\noindent \textbf{Interaction Effects:} Mask M has a strong effect on performance when combined with specific positional PE or structural embeddings E. For example,
% Mask-Sparsity and Positional Embeddings exhibit an interaction effect of $\eta^2=0.05$ on structure, with 
TPE consistently outperforms CPE when using masks (e.g., T2 with a mask). Furthermore, there is a notable interaction between PE and E ($\eta^2\in[0.19, 0.26]$). 
% These findings highlight the importance of integrating sparse attention masks in tabular models.

\noindent \textbf{Non-significant Factors:} Special tokens (T) and Bias (B) do not significantly impact performance, either alone or in interactions (except Bias with Mask; see Appendix \ref{appendix:bias_and_sparse_masks}). This suggests that token structure encoding is less critical compared to sparsity mechanisms or positional embeddings, aligning with prior work emphasizing the role of better structure-aware representations.

\begin{table}[thb]
\centering
\caption{ANOVA results showing the effect sizes ($\eta^2$) and p-values for each factor on performance across different tasks (In Domain, Structure, Consistency, Compositional, and Mixability). Significant effects are indicated with bold text if (p $\leq$ 0.05).}
\setlength{\tabcolsep}{2pt}
\resizebox{0.5\textwidth}{!}{%
\begin{tabular}{p{1.3cm}cccc}
\hline
\textbf{Factor} & \multicolumn{1}{c}{\textbf{In Domain}} & \multicolumn{1}{c}{\textbf{Structure}} & \multicolumn{1}{c}{\textbf{Consistency}} & \multicolumn{1}{c}{\textbf{Compositional}} \\
                & {\small \textbf{$\eta^2$ (p-value)}}  & {\small \textbf{$\eta^2$ (p-value)}}  & {\small \textbf{$\eta^2$ (p-value)}}  & {\small \textbf{$\eta^2$ (p-value)}} \\
\hline
T        &     0.00 &      0.00 &       0.00 &          0.00 \\
M        &     0.04 &      0.07 &       0.01 &          0.00 \\
PE       &     \textbf{0.19 }&      \textbf{0.27} &       \textbf{0.19} &          \textbf{0.26} \\
B        &     0.01 &      0.01 &       0.00 &          0.00 \\
E        &     \textbf{0.20} &      \textbf{0.15} &       \textbf{0.30} &          \textbf{0.26} \\
TM      &     0.00 &      0.01 &       0.01 &          0.02 \\
T$\times$PE     &     0.00 &      0.00 &       0.00 &          0.01 \\
T$\times$B      &     0.00 &      0.00 &       0.00 &          0.00 \\
T$\times$E      &     0.00 &      0.00 &       0.00 &          0.00 \\
M$\times$PE     &     \textbf{0.08} &      \textbf{0.05} &       \textbf{0.07} &          \textbf{0.04} \\
M$\times$B      &     0.02 &      \textbf{0.04} &       0.02 &          0.01 \\
M$\times$E      &     0.09 &      0.04 &       0.07 &          0.04 \\
PE$\times$B     &     0.01 &      0.00 &       0.00 &          0.00 \\
PE$\times$E     &     \textbf{0.19} &      \textbf{0.21} &       \textbf{0.18} &          \textbf{0.26} \\
B$\times$E      &     0.01 &      0.01 &       0.01 &          0.00 \\
\hline
\end{tabular}%
}
\label{tab:exp:anova}
\end{table}

\subsection{Impact of Structural Encoding: Performance Differences Across Models}

\begin{figure}[t]
  \centering\includegraphics[width=0.48\textwidth]{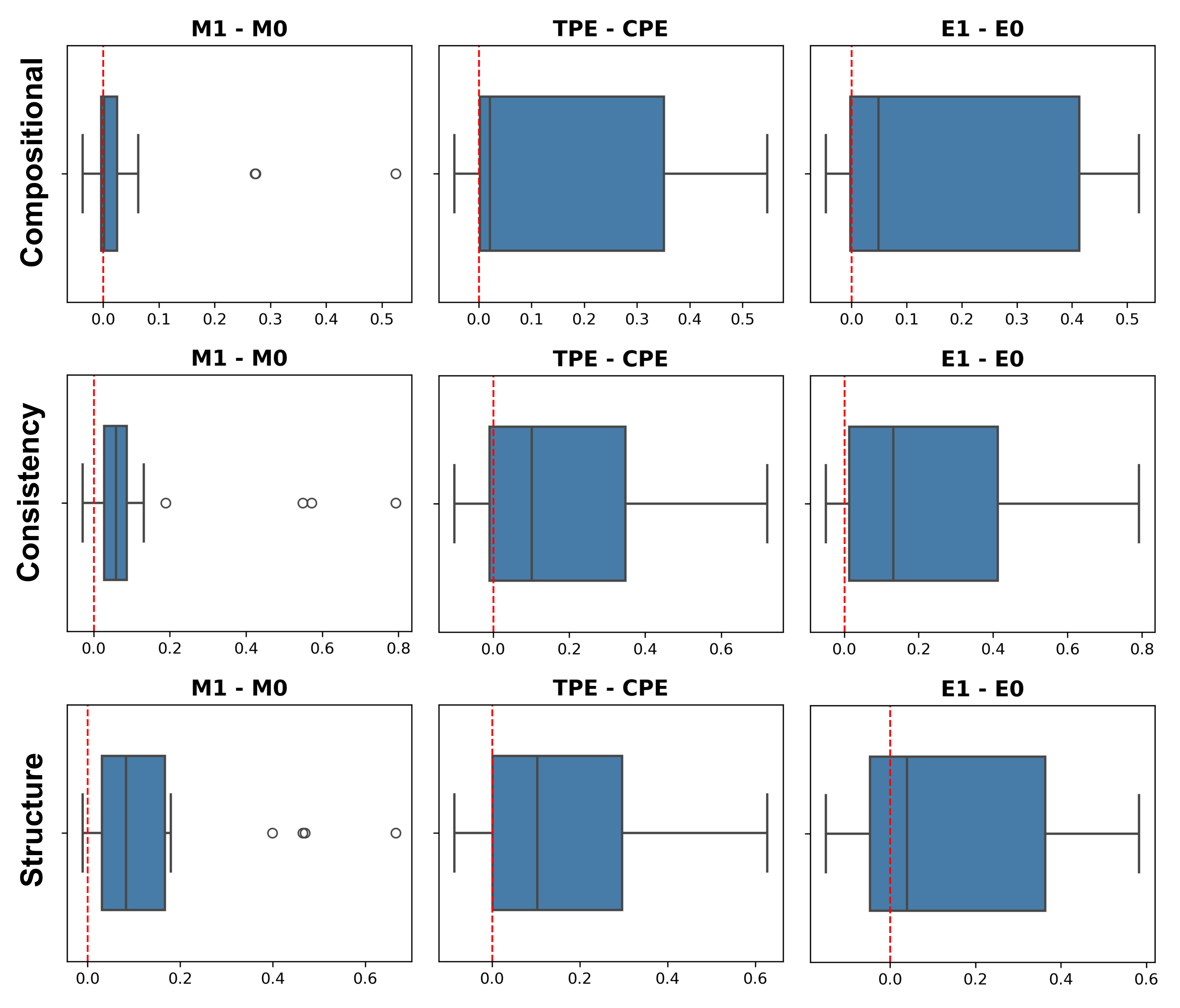}
 \caption{Denotation Accuracy differences between two structural encoding components (left $-$ right) while keeping all other factors unchanged.}
\label{fig:diff}
\end{figure}

We analyze further key ANOVA findings by focusing on the most significant factors (Figure \ref{fig:diff}). The figure illustrates performance differences between two treatments of a factor (e.g., PE), while keeping other factors (e.g., T, E, M, B) constant. First, sparse masking M1 consistently improves performance over M0. This supports the hypothesis that restricting cell attention in table encoding mitigates spurious correlations and enhances generalization. Second, absolute encoding generally yields better generalization than relative encoding (TPE $>$ CPE and $E1>E0$). Although counterintuitive, since relative encodings should help capturing table invariance (e.g. swapping two rows or columns), this is likely due to the task nature, where SQL query interactions with table during question answering require absolute embeddings to capture rule-based relationships effectively. We conduct the same experiments for tokens and bias, with results provided in Appendix \ref{appendix:impact_of_structure_embeddings}.

\subsection{Validating Structural Encoding Insights: Consistency Across Synthetic and Real-World Data}

To validate our findings on real-world data, we evaluated the WikiSQL dataset, testing 120 model configurations (see Table \ref{fig:wsql}).

\noindent \textbf{Performance of M1:} Overall, the M1 mask outperforms the baseline (M0) across all configurations, demonstrating that introducing a sparse mask significantly enhances generalization. More experiments on M1 can be find in Appendix \ref{appendix:mask_sparsity_effect}.

\noindent \textbf{Performance of Other Masks:} While M1 offers a good trade-off between sparsity and performance, other masks such as M5$^{\textcolor{red}{*}}$ or M3, despite being significantly sparser (see Figure \ref{appendix:sparsemask} in the appendix), still outperform M0 by a wide margin and achieve comparable performance to M1. Notably, these masks are at least twice as sparse as M1, with M6 having 99\% of its attention masked.

\noindent \textbf{Absolute vs. Relative Encoding:}  Some configurations failed to converge due to overfitting, particularly when absolute information  is absent (CPE, E0). Without absolute positional cues, the model struggles to differentiate cells, leading to poor generalization. This aligns with our ANOVA analysis on synthetic data -- see Appendix figure \ref{appendix:alignment} for more details.

\begin{table*}[t]

\centering
\caption{This table summarizes all possible model combinations from the literature (T, E, PE, M, B) as well as our proposed configurations (M2, ..., M6, and T2) evaluated on the WikiSQL dataset.}

\resizebox{\textwidth}{!}{%
\begin{tabular}{lll|cccccccccccccc}
\hline
\multirow{2}{*}{PE} & \multirow{2}{*}{E1} & \multirow{2}{*}{T} & \multicolumn{2}{c}{M0} & \multicolumn{2}{c}{M1} & \multicolumn{2}{c}{M2} & \multicolumn{2}{c}{M3} & \multicolumn{2}{c}{M4\textcolor{red}{*}} & \multicolumn{2}{c}{M5\textcolor{red}{*}} & \multicolumn{2}{c}{M6\textcolor{red}{*}} \\
& & & B0 & B1 & B0 & B1 & B0 & B1 & B0 & B1 & B0 & B1 & B0 & B1 & B0 & B1 \\
\hline
\multirow{6}{*}{CPE} & \multirow{3}{*}{E0} & T2 & 26.1 & 38.6 & 81.3 & 81.6 & 30.2 & 29.7 & 58.0 & 56.3 & 29.9 & 29.9 & 61.4 & 57.5 & 29.2 & 28.8 \\
& & T1 & 26.1 & 26.6 & 82.3 & 81.5 & 30.0 & 30.0 & 57.1 & 54.7 &  &  &  &  &  &  \\
& & T0 & 22.7 & 23.1 & 79.6 & 76.8 & 29.8 & 29.5 & 36.5 & 34.4 &  &  &  &  &  &  \\
& \multirow{3}{*}{E1} & T2 & 75.7 & 73.0 & 82.1 & 81.6 & 78.5 & 77.6 & 81.8 & 81.2 & 79.2 & 78.6 & 81.9 & 81.2 & 29.3 & 29.6 \\
& & T1 & 79.0 & 79.0 & 82.2 & 81.5 & 78.5 & 78.0 & 82.4 & 81.4 &  &  &  &  &  &  \\
& & T0 & 29.8 & 28.8 & 79.2 & 78.3 & 77.2 & 78.0 & 71.0 & 66.2 &  &  &  &  &  &  \\
\hline
\multirow{6}{*}{TPE} & \multirow{3}{*}{E0} & T2 & 79.4 & 79.6 & 82.0 & 82.0 & 78.9 & 78.2 & 78.7 & 79.1 & 80.1 & 80.0 & 79.0 & 78.5 & 80.5 & 80.4 \\
& & T1 & 79.5 & 79.6 & 81.9 & 82.4 & 79.3 & 78.8 & 79.2 & 79.5 &  &  &  &  &  &  \\
& & T0 & 70.6 & 71.4 & 82.5 & 82.5 & 79.3 & 78.9 & 75.3 & 73.9 &  &  &  &  &  &  \\
& \multirow{3}{*}{E1} & T2 & 79.4 & 79.3 & 82.2 & 82.3 & 78.6 & 78.0 & 79.8 & 79.4 & 80.5 & 80.5 & 79.7 & 79.3 & 80.4 & 79.8 \\
& & T1 & 79.6 & 79.7 & 81.9 & 81.3 & 79.0 & 78.5 & 79.1 & 79.9 &  &  &  &  &  &  \\
& & T0 & 77.4 & 77.5 & 82.6 & 82.6 & 79.2 & 79.1 & 77.2 & 76.8 &  &  &  &  &  &  \\
\hline
\end{tabular}%
}
\label{fig:wsql}
\end{table*}

\subsection{M3: An Efficient Sparse Attention Mask for Faster and Improved Performance}

In this section, we focus on M3 which has a very good performance overall, along with a high sparsity level, and study its impact on model efficiency.
%We established that M1 masks enhance performance and that other sparse masks yield similar benefits. 
% Beyond generalization, sparsity can significantly improves computational efficiency \cite{10.1145/3530811}. 
For this experiment, we use \href{https://github.com/pytorch-labs/attention-gym}{FlexAttention}  and FlashAttention (FA2) \cite{dao2022flashattention}, both optimized for sparse matrix operations. FlexAttention, a PyTorch module specifically designed for efficient sparse attention computation, performs optimally with block-based sparse attention masks, aligning well with the structure of M3 (see Figure \ref{appendix:sparsemask} in appendix).
The results are presented in Figure \ref{fig:efficient}. The cumulative distribution of WikiTableQuestions (blue curve) reveals that many tables exceed 1024 tokens, the common limit in fine-tuned models, restricting encoding to 80\% of WikiTableQuestions. 
Sparse masks mitigate this limitation.
We assessed Forward and Backward Speedup using M3 across sequence lengths of 1024, 2048, 4096, 8192, and 16,384 tokens. Speedup is computed as the ratio of the accelerated masked attention (either Flex or Flash) to the standard PyTorch Sdpa attention. 
Up to 4096 tokens, FA2 is the most efficient, achieving a 2$\times$ speedup over standard attention at 2048 tokens. For longer sequences, FlexAttention reach a 50$\times$ forward speedup and (16$\times$ for backward) when encoding tables of 16,384 tokens.

\begin{figure}[tbh]
  \centering
  \includegraphics[width=0.5\textwidth]{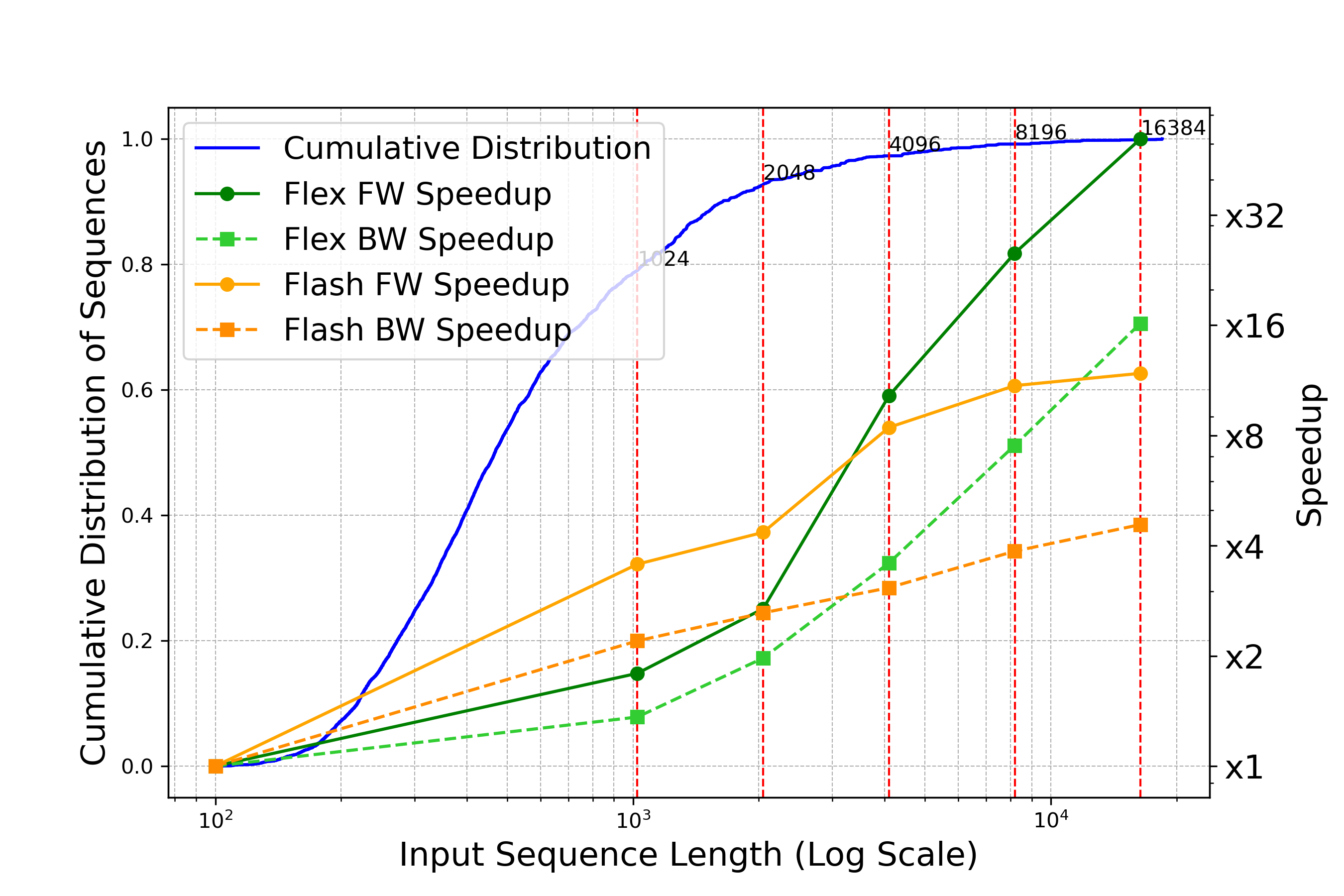}
\caption{Cumulative Distribution of Sequence Lengths and Relative Computation Speedup: The primary y-axis (left) represents the cumulative distribution of sequence lengths in log scale, while the secondary y-axis (right) shows the relative computation speedup for FlexAttention and FlashAttention2 across different sequence lengths in x-axis. } 
\label{fig:efficient}
\end{figure}

\begin{table}[t]
\centering
\caption{Comparison of Model Performance on the Synthetic dataset. "ALL" represents the average score across Structure, Compositionality, and Robustness, alongside WikiSQL test data.}
\small
\begin{tabular}{lcc}
\hline
\textbf{Model} & \textbf{ALL} & \textbf{WSQL} \\
\hline
\multicolumn{3}{@{}c@{}}{\textit{Literature Models}} \\

MATE \cite{eisenschlos2021mate} & 79.2 &  82.6 \\
TableFormer \cite{yang2022tableformer} & 23.1 & 60.5 \\
TAPAS \cite{herzig2020tapas} & 77.4 &  78.0 \\
TAPEX \cite{liu2021tapex} & 79.5 &  74.7 \\

\multicolumn{3}{@{}c@{}}{\textit{Our models}} \\
T2 M0 TPE B E1  &79.3 &  78.5 \\
T2 M3 TPE B E1  & 79.4 &  80.3 \\

\hline
\end{tabular}
\label{tab:M3}
\end{table}

Having established M3 computational efficiency, we now evaluate its empirical performance on synthetic and real datasets, comparing against literature baselines and our own implementations (see Table \ref{tab:M3}).
From Table \ref{tab:M3}, we observe that the M3 mask performs well on the ALL dataset, which represents the average results across all our synthetic datasets, and remains competitive on WikiSQL. This highlights its effectiveness as a candidate for ultra-sparse table encoding. Additionally, compared to existing models, our M3 implementation maintains strong performance while offering significant computational advantages, making it a promising solution for scalable table-based question-answering tasks.

\subsection{Enhancing Generalization with Sparse Attention: The Impact of Non-Random Table Structures}

We have identified key factors for encoding tables, and now we explore another critical element for generalization, namely table structure determinism. In datasets from the literature, the number of unique tables is relatively low. For instance, WikiTableQuestions contains only 2k unique tables in a training set of 11k examples. To investigate the impact of deterministic table structures, we compared model performance when trained on either fully deterministic or randomized data.

\begin{figure}[t]
  \centering
  \includegraphics[width=0.48\textwidth]{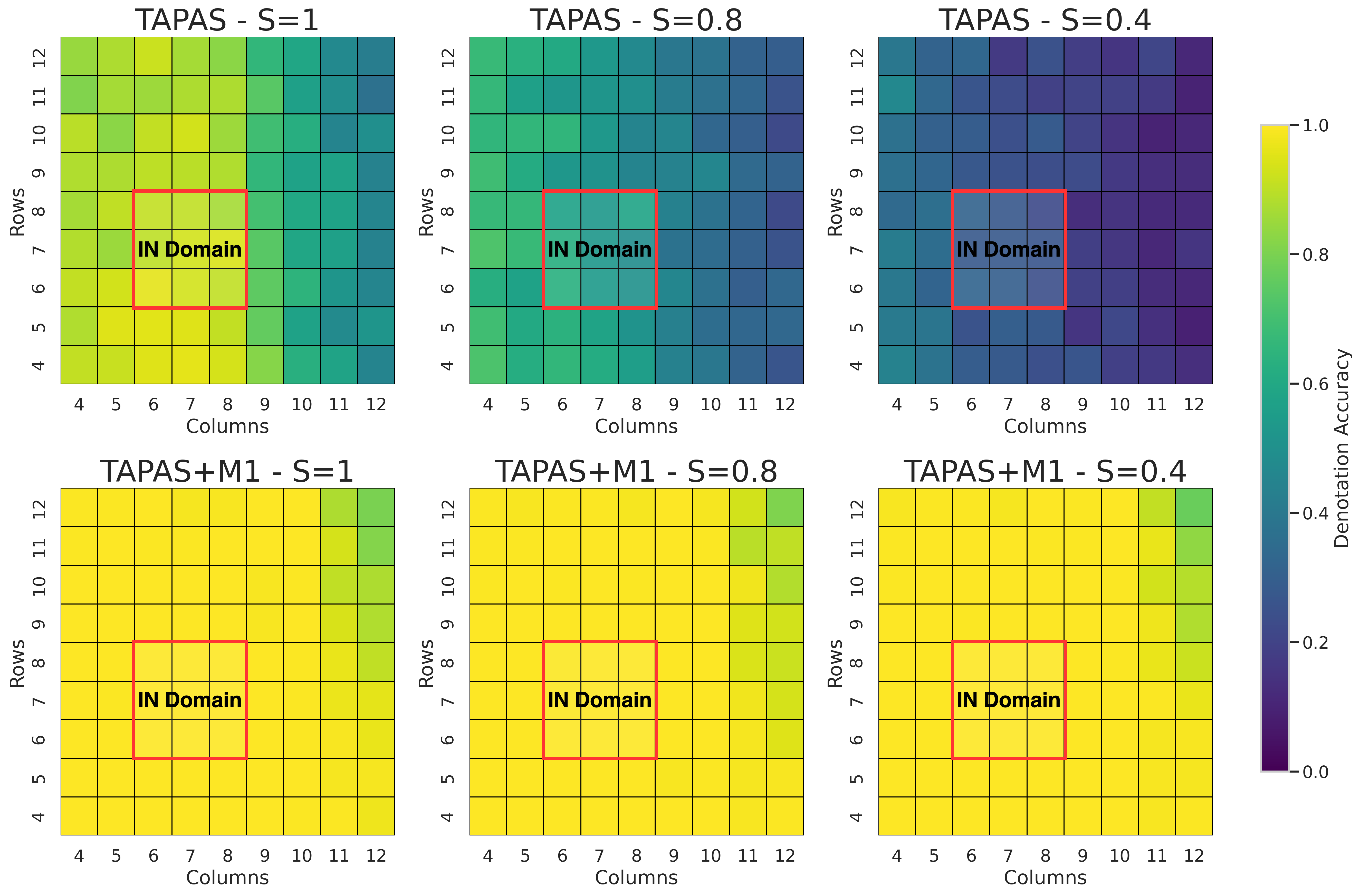}
\caption{Results for TAPAS and TAPAS+M1 under varying Mixability levels (S). All models have been trained on data with S=1, where the transition matrix for table creation is fully deterministic, and tested on increasingly challenging similarity levels, down to S=0, where the transition matrix is uniformly random. For this experiment, we exclusively used the “SELECT cx WHERE cy = vy” template SQL query.}
\label{fig:cpeoverfit}
\end{figure}

Figure \ref{fig:cpeoverfit} compares the performance of two models, TAPAS and TAPAS+M1, across varying levels of mixability level (section \ref{sec:mixability}). The top row shows TAPAS results, while the bottom row displays TAPAS+M1 results. Both models are trained on data with maximum mixability (S=1) and a restricted number of rows and columns, and tested on datasets with decreasing mixability (0.8 to 0) and larger tables. 

We observe that TAPAS exhibits significant overfitting, with performance declining both “In Domain” (outlined in red) and “Out of Domain” (larger tables).
In contrast, the TAPAS+M1 model (bottom row) demonstrates resilience to these shifts. Even at lower mixability levels, TAPAS+M1 maintains high and stable accuracy, indicating that the sparse mask (M1) helps mitigate overfitting and improves generalization to out-of-domain data.

\section{Conclusion and Recommendations}

We conducted a comprehensive analysis of encoding techniques for data tables, highlighting the importance of sparse attention and absolute positional encoding in model generalization. Our study systematically evaluated table encoding components, as well as their interactions, providing insights for improving table representation in Transformer-based architectures.

Our findings show that sparse attention masks reduce spurious correlations and enhance structural representation. In particular, we propose the mask M3 that achieves a good efficiency-effectiveness trade-off. Additionally, we show that absolute positional encoding (TPE, E1) is essential, as models relying solely on relative encoding struggle with generalization.
Beyond accuracy, sparse masks like M3 achieve up to 50× forward and 16× backward speedup for large tables, enabling efficient scaling for real-world applications.
% We also highlight table structure determinism as a key factor, where sparse attention helps mitigate overfitting and improves robustness in out-of-domain settings.

Overall, our work provides a foundation for efficient and scalable  transformers for information extraction from tables of data. High-quality encodings have proven essential when integrating these representations into LLMs, particularly in domains like visually rich document understanding~\cite{ee-mllm,pix2struct}. %We argue that this is also the case for tables, and future works will investigate how to use these encoders as an input for LLMs. 
Future works will investigate how to leverage our findings in decoder-only LLM architectures. %use these encoders as an input for LLMs. 

%Dans cette étude nous proposons de compléter chaque composants permettant 

%In this study, we explored a set of techniques for encoding tabular structures and proposed various sparse mask configurations. 

% Our findings provide novel insights into the impact of encoding choices on model performance. 
%The key recommendations for future work are as follows. Using sparse attention, especially the M1 mask, improves generalization. Structural embeddings may not be the best option for preserving table structure, as other methods perform better. Table Positional Embeddings (TPE) are preferred to avoid overfitting compared to Cell Positional Embeddings (CPE). Not using structural tokens, or using T2, with sparse masks, are sufficient for capturing table structure. Combining the M1 mask with bias leads to strong model performance. 
% These recommendations are particularly useful for tasks involving logical reasoning on tabular data.\\
% In conclusion, our findings provide valuable insights into improving the generalization of transformer models on tabular data and offer a foundation for future research aimed at enhancing generalization across diverse table structures encoding.

\newpage

\section{Limitations}

Our study highlights the effectiveness of absolute table encoding for generalization in question-answering tasks. While relative encoding is often expected to enhance generalization, current approaches may not fully capture rule-based relationships in SQL queries. This opens an exciting direction for future work to develop improved relative encoding mechanisms that better integrate structural rules.

Additionally, while our experiments primarily rely on synthetic datasets, we include preliminary evaluations on real-world data (WTQ and WSQL). Expanding real-world benchmarks will further strengthen our findings and enhance applicability across diverse scenarios.

Lastly, our analysis focuses on the BART model as a baseline. Extending this work to other encoder-decoder architectures and decoder-only models presents an exciting opportunity for future research, potentially extending the impact of our approach.

\bibliography{main}

\appendix

\section{Reproductibility statement}

All experiments in this paper were conducted using publicly available datasets and open-source libraries, including \cite{pasupat2015compositional, shi2020potential, zhong2017seq2sql}. Detailed descriptions of model architectures, hyperparameters, and training configurations are provided in the section \ref{section:training_pipeline}. To implement our experiments, we used the transformer library \cite{wolf2019huggingface}. The code will be published upon publication. All experiments were conducted on work stations with 80GB NVIDIA A100, 16 or 32GB NVIDIA V100 SXM2-HBM2 GPUs. %cite experimaestro ?

\section{Methodology Details}

\begin{table}[h!]
\centering
\begin{tabular}{|l|l|}
\hline
\textbf{Parameter}                    & \textbf{Values}                                                            \\ \hline
\multirow{3}{*}{Input Token Structures}  & T1  \\ \cline{2-2} 
                                        & T2  \\ \cline{2-2} 
                                        & T0   \\ \hline
\multirow{8}{*}{Mask Sparsity Levels}    & M0   \\ \cline{2-2} 
                                        & M1   \\ \cline{2-2} 
                                        & M2   \\ \cline{2-2} 
                                        & M3   \\ \cline{2-2} 
                                        & M4   \\ \cline{2-2} 
                                        & M5   \\ \cline{2-2} 
                                        & M6   \\  \hline
\multirow{2}{*}{Positional Embeddings}   & TPE  \\ \cline{2-2} 
                                        & CPE  \\ \hline
\multirow{2}{*}{Encoding Structure Bias} & Bias \\ \cline{2-2} 
                                        & No Bias \\ \hline
\multirow{2}{*}{Tabular Structure Embeddings} & E1  \\ \cline{2-2} 
                                              & E0 \\ \hline
\end{tabular}
\caption{Overview of the experimental parameters used in our study. These include  input token structures, mask sparsity levels, positional embeddings, encoding structure biases, and tabular structure embeddings.}
\end{table}

\begin{table}[t]
\centering
\caption{State-of-the-art models and structural encoding methods.}
\small
\begin{tabular}{lccccc}
\hline
\textbf{Model} & T & E1 & PE & B & M \\
\hline
MATE & T0 & E1 & CPE & B0 & M1\\
TableFormer & T0 & E0 & CPE & B0 & M0 \\
TAPAS & T0 & E1 & TPE & B0 & M0 \\
TAPEX & T1 & E0 & TPE &B0 & M0 \\
\hline
\end{tabular}
\label{tab:charac}
\end{table}

\subsection{TableFormer Bias}
\label{appendix:tableformer_bias}
 In TABLEFORMER, attention biases are designed to handle various table-text structural relationships, using 13 types of biases to capture row, column, header, and sentence relationships in tabular data. These biases include mechanisms for recognizing same row and column information, linking cells to their respective headers, and incorporating sentence-to-cell grounding to enhance the understanding of tables in context. Each bias type is then  associated with a learnable scalar. For more details, see \cite{yang2022tableformer}.

\subsection{SQL Query for training.}
\label{appendix:sql_train}
The following SQL query templates were used to generate synthetic data for our experiments. Each template includes various possible choices for conditions, making them versatile and applicable to different table structures:

\begin{itemize}
    \item \texttt{SELECT cx WHERE cy =|!= vy AND|OR cz =|!= vz AND|OR cw =|!= vw AND|OR cl =|!= vl}
    \item \texttt{SELECT cx WHERE cy =|!= vy AND|OR cz =|!= vz AND|OR cw =|!= vw}
    \item \texttt{SELECT cx WHERE cy =|!= vy AND|OR cz =|!= vz}
    \item \texttt{SELECT cx WHERE cy =|!= vy}
    \item \texttt{SELECT cx}
    \item \texttt{SELECT cx LIMIT k}, where $k \in \{1, 2, 3\}$
    \item \texttt{SELECT cx WHERE cy = (SELECT cy WHERE cy = vy)}
    \item \texttt{SELECT cx WHERE cy IN (vy | vy, vy | vy, vy, vy)}
\end{itemize}

\subsection{Generating tables for mixability}
\label{sec:mixability-generation}
To generate tables where cell content is more or less determined by the previous cells in the row, we use two base transition matrices: a deterministic transition matrix $M^{\text{deter}}$, and a uniform matrix $M^{\text{unif}}$, where transitions are equally probable. The transition matrix for our experiments is a weighted combination of these two matrices:
$$M^{\text{transi}} = S \cdot M^{\text{deter}} + (1 - S) \cdot M^{\text{unif}}$$
During training, we set $S=1$ for structured tables, while during testing, we use $S=0$ to introduce randomness, simulating real-world data variability.

\section{Additional Results}

\subsection{Performance Differences Across Models}
\label{appendix:performance_differences_across_models}

\begin{figure}[t]
  \centering\includegraphics[width=0.48\textwidth]{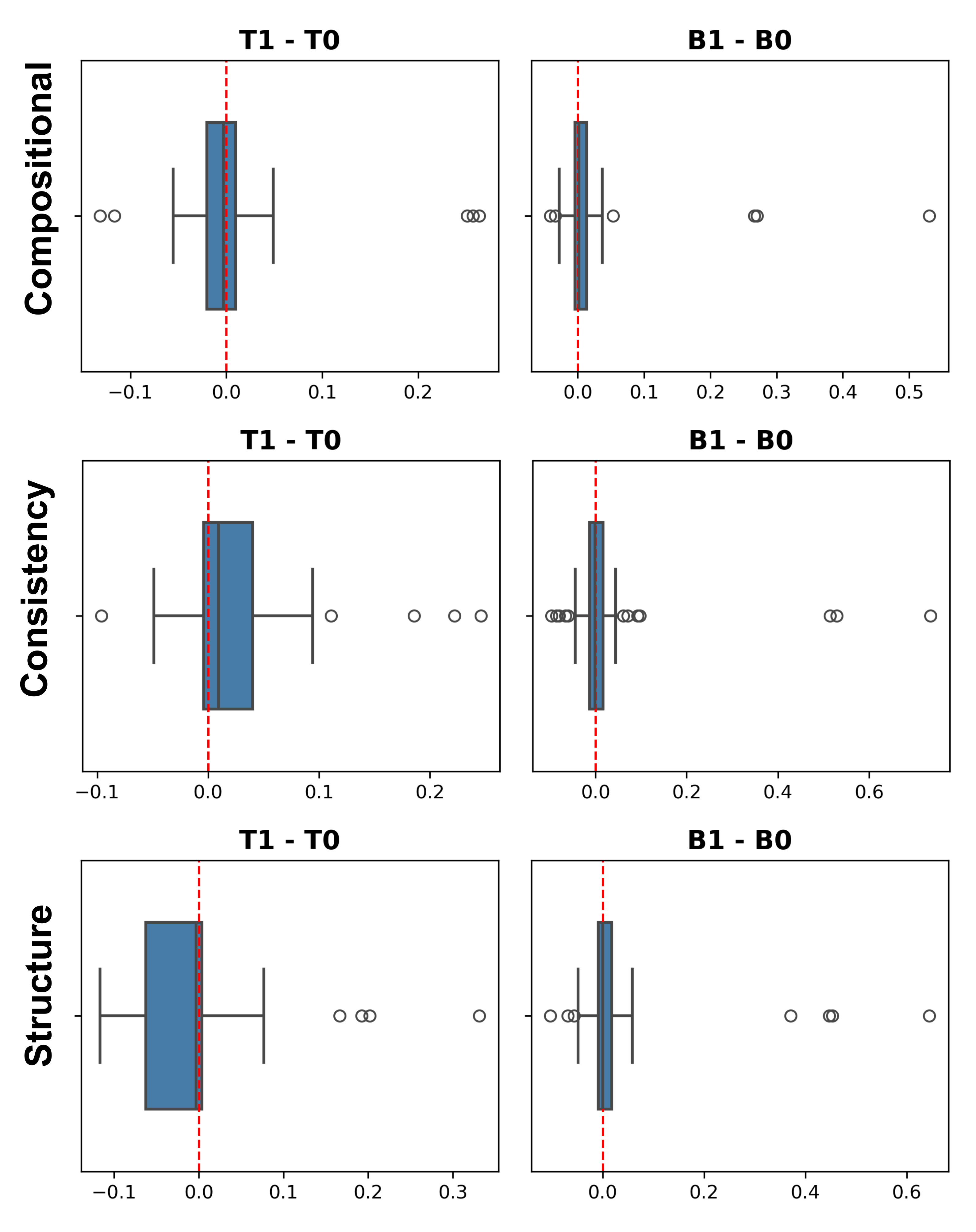}
 \caption{This figure highlights the differences between two structural encoding components while keeping all other factors unchanged.}
\label{fig:diff2}
\end{figure}

We report in Figure \ref{fig:diff2} the effect of special tokens (T) and bias (B) on the three synthetic datasets.

\subsection{Mask Sparsity Effects}
\label{appendix:mask_sparsity_effect}

Based on the ANOVA results and the strong significance of mask sparsity on accuracy variation, we selected compositions of factors from the literature to evaluate against the sparse mask M1, across both generated and real-world datasets (WikiSQL, WikiTableQuestions). M1 was tested individually because it had previously demonstrated the best overall performance in evaluations. As shown in Table \ref{tab:literature}, the results demonstrate that adding the M1 mask leads to substantial performance improvements in both in-domain, out-of-domain generalization tasks and real-world-datasets.

Across all out-domain data (Structure, Robustness, Compositional, Mixability), the addition of the M1 mask yields significant gains, with improvements reaching over 10\%. For example, the T0/M1 model outperforms the T0/M0 model by 9.7 points in robustness and 5.6 points in compositional generalization. This indicates that sparse masks help the models generalize better.

For real datasets, the sparse mask again improves performance. The T0/M1 model achieves 46.8\% on WikiSQL, compared to 34.6\% for the T0/M0 model, representing a significant 12.2-point gain. This underscores the practical benefit of sparse masks in real-world applications, where data distributions can be more variable and noisy.

Overall, this experiment highlights the critical role of mask-based sparsity (M1) in improving model performance across a wide range of tasks and datasets. The results not only confirm the hypothesis that sparse masks enhance generalization but also show that these gains extend across both synthetic and real-world datasets, making them a valuable addition to existing models in the literature.

\begin{table*}[tbh]
\centering
\caption{Comparison of model performance on generalization test sets, highlighting the effect of mask M1 across both in-domain and out-of-domain tasks (Structure (Struc), Mixability (Mixab), Consistency (Consi),  Compositional (Comp) and real datasets (WTQ, WSQL). Models with mask M1 generally show improved accuracy, particularly in robustness and compositional generalization, compared to models without the mask (M0). This demonstrates the impact of mask-based sparsity on handling diverse data distributions.}
\label{tab:gen_test_results}
\small
\setlength{\tabcolsep}{4.2pt} % Adjust space between columns
\renewcommand{\arraystretch}{1.2} % Increase row height for better readability
\begin{tabular}{|l|ccccc|cc|}
\hline
{\small \textbf{Model}} & {\small \textbf{InDomain}}& {\small \textbf{Mixab}} & {\small \textbf{Consi}}  & {\small \textbf{Comp}} & {\small \textbf{Struct}}   & {\small \textbf{WSQL}}  & {\small \textbf{WTQ}}  \\
\hline
{\small \textbf{T0/M0/TPE/B0/$\text{E1}_{\textbf{tapas}}$}}   & 99.8 &  98.9 &   76.3 & 62.8 &  66.3 &   84.8 &  34.6 \\
{\small \textbf{T0/M1/TPE/B0/E1}}    & 99.9 &  98.8 &   88.4 &  62.8 &  76.0 &   87.6 &  46.8 \\
{\small \textbf{T1/M0/TPE/B0/$\text{E0}_{\textbf{tapex}}$}} & 98.4 &  98.9 &   71.3 &  62.9 &  66.3 &    87.1 &  52.4 \\
{\small \textbf{T1/M1/TPE/B0/E0}} & 99.8 &  99.1 &   90.2 &  62.6 &  71.5 &    86.0 &  51.4 \\
{\small \textbf{T0/M0/TPE/B1/$\text{E0}_{\textbf{tableformer}}$}}   & 99.9 &  98.5 &   80.2 &  62.7 &  78.0 &   83.3 &  42.4 \\
{\small \textbf{T0/M1/TPE/B1/E0}}   & 99.8 &  99.2 &   86.2 &  62.4 &  81.1 &  86.9 &  46.5 \\

\hline
\end{tabular}
\label{tab:literature}
\end{table*}

\subsection{Alignment Between Synthetic and Real Data:}
\label{appendix:alignment}
\begin{figure}[t]
  \centering
  \includegraphics[width=0.5\textwidth]{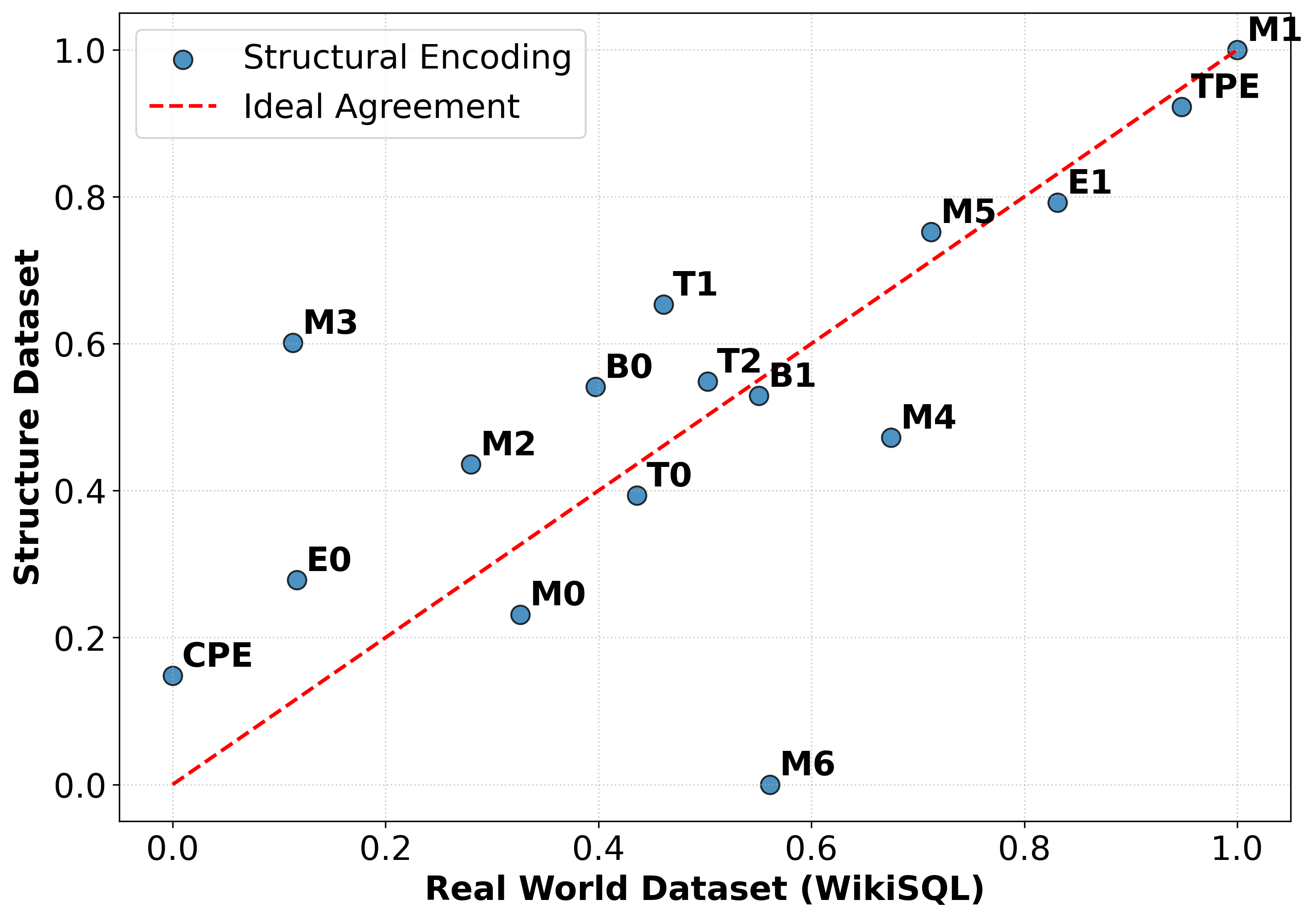}
\caption{This figure highlights the agreement between our "structure" synthetic dataset (x-axis) and the real-world WikiSQL dataset (y-axis) -- with normalized (max) metrics, and using the mean over all experiments with a given table encoding component. The diagonal red line represents the ideal agreement where identical results are obtained on both datasets.}
\label{appendix:fig:agreement}
\end{figure}

Figure \ref{appendix:fig:agreement} assesses consistency across datasets. Red markers indicate perfect agreement, with standardized means comparing encoding effects across synthetic and real data. A strong correlation confirms alignment, except for M6, where extreme sparsity prevents the model from recognizing table vocabulary, degrading performance.

\subsection{Bias and Sparse Masks.}
\label{appendix:bias_and_sparse_masks}

As the ANOVA results show (see Table \ref{tab:exp:anova}), the interaction between bias and mask has a significant effect on structure. To analyze this interaction effect, we average the bias and mask results across interaction configurations, as shown in Table \ref{tab:biasxmask}. We observe that the models that benefit the most from the addition of bias are those using no mask (M0). For all other masks, adding bias has only a minor effect.

% While bias alone improves accuracy, its combination with mask sparsity enhances both the stability and generalization across structural generalization tasks.
% 

%As illustrated in Table \ref{tab:biasxmask}, models \emph{without} bias (B0) generally perform worse, with lower mean accuracy and higher variance across almost all sparsity levels, especially at the extremes (M0 and M6). 
% For example, at M0 (no sparsity), models without bias achieve a mean accuracy of 48.6 and a variance of 0.06, whereas models with bias significantly improve with a mean accuracy of 66.5 and a variance of only 0.01. 
% Interestingly, the only exception is with the M1 mask, where models without bias perform nearly as well as those with bias, but the gain remains under 1 point.

T%hese findings highlight the importance of incorporating both bias and mask sparsity in structured table representation models. While bias alone enhances robustness, the combination of bias and sparse masks offers a clear advantage in stabilizing performance and improving generalization. These conclusions are further supported by Table \ref{tab:literature}, which compares the performance of TableFormer (a biased model) with and without sparse masks across all datasets tested in this paper.

\begin{table}[b]
\centering
\caption{Performance (Structured generalization) with and without Bias for different sparsity levels.}
\label{tab:system_performance}
\small % Reduce font size for the entire table
% Visionner le code LaTeX du paragraphe 0

\begin{tabular}{|l|c|c|c|c|}
\cline{2-5}
\multicolumn{1}{l|}{} & \multicolumn{2}{c|}{B0} & \multicolumn{2}{c|}{B1}\tabularnewline
\hline 
\textbf{Sparsity}  & \textbf{Mean}  & \textbf{Var} & \textbf{Mean}  & \textbf{Var}\tabularnewline
M0           &      59.1 &  0.019 &  69.6 & 0.004 \\
M1           &      75.5 &   0.001 & 75.7 & 0.001 \\
M2           &      71.4 &   0.002 & 70.6 & 0.003 \\
M3           &      67.5 &   0.000 & 66.9 & 0.001 \\
M4$^{\textcolor{red}{*}}$           &      71.4 &  0.008 & 72.8 &  0.002 \\
M5$^{\textcolor{red}{*}}$           &      75.6 &   0.000 & 74.2 & 0.000 \\
M6$^{\textcolor{red}{*}}$           &      71.5 &  0.000 &  72.0 &  0.001 \\
\hline 
\end{tabular}

\label{tab:biasxmask}
\end{table}

\subsection{Impact of Structure Embeddings}
\label{appendix:impact_of_structure_embeddings}

\begin{figure}[tbh]
  \centering
  \includegraphics[width=0.4\textwidth]{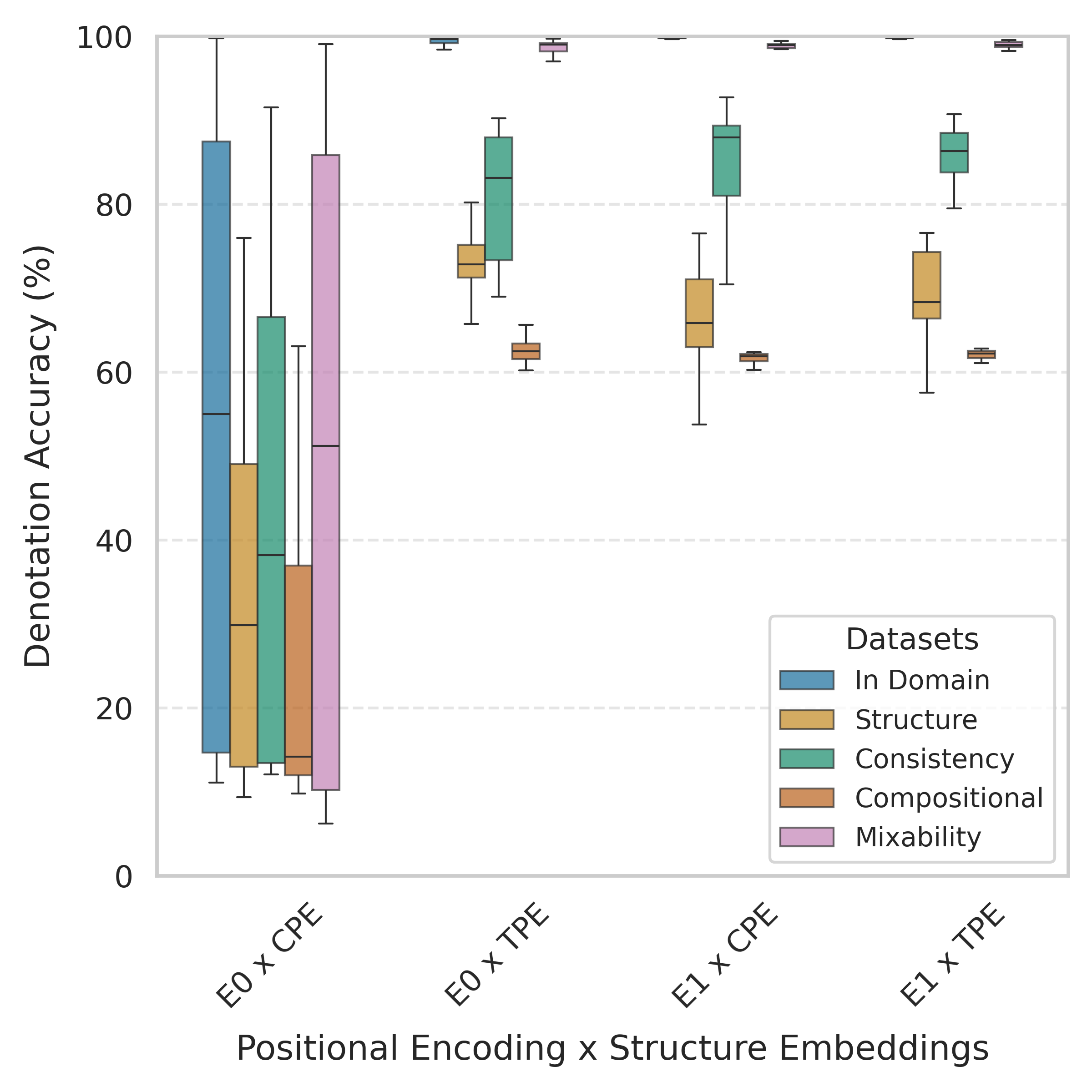}
\caption{Impact of Positional Encoding (CPE, TPE) in Combination with Structure Embeddings (RCE, NRCE). } 
\label{fig:StructEmbeddingxpositional}
\end{figure}

We tested the combined effect of structure embedding and positional encoding across multiple datasets, as shown in Figure \ref{fig:StructEmbeddingxpositional}. The results clearly demonstrate that models using CPE without E1 struggle to converge -- most probably because there is no more absolute positioning information. 
% In contrast, models incorporting E1 with CPE show significantly more robust performance across all datasets. 
%
However, using standard positional embeddings (TPE) demonstrates better performance on the structure and compositional datasets, showing that table information can be injected by with other factors such as masking and bias.

\section{Sparse Mask Details}

\begin{figure*}[ht]
  \centering
  \includegraphics[width=1\textwidth]{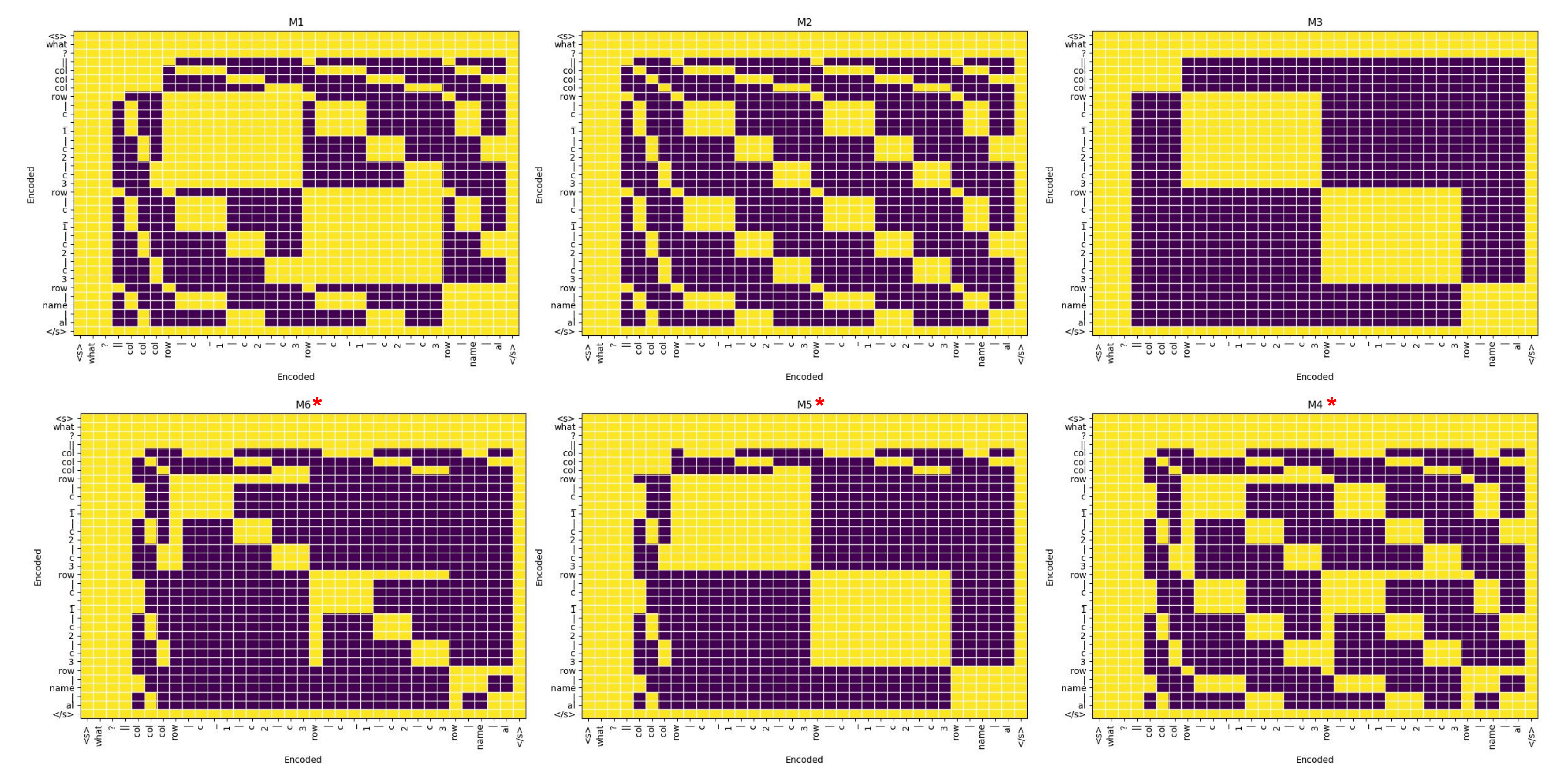}
  \caption{Visualization of sparse masks ranging from M0 (no sparsity) to M6\textcolor{red}{*} (high sparsity) for T2 structural token. As the mask sparsity level increases from M0 to M6, the sparsity of the mask increase. Masks marked with a red star \textcolor{red}{*}, such as M6\textcolor{red}{*}, indicate that they are only applicable to special tokens corresponding to Row-Column Cells (T2).}
  \label{appendix:sparsemask}
\end{figure*}

\begin{figure*}[ht]
  \centering
  \includegraphics[width=1\textwidth]{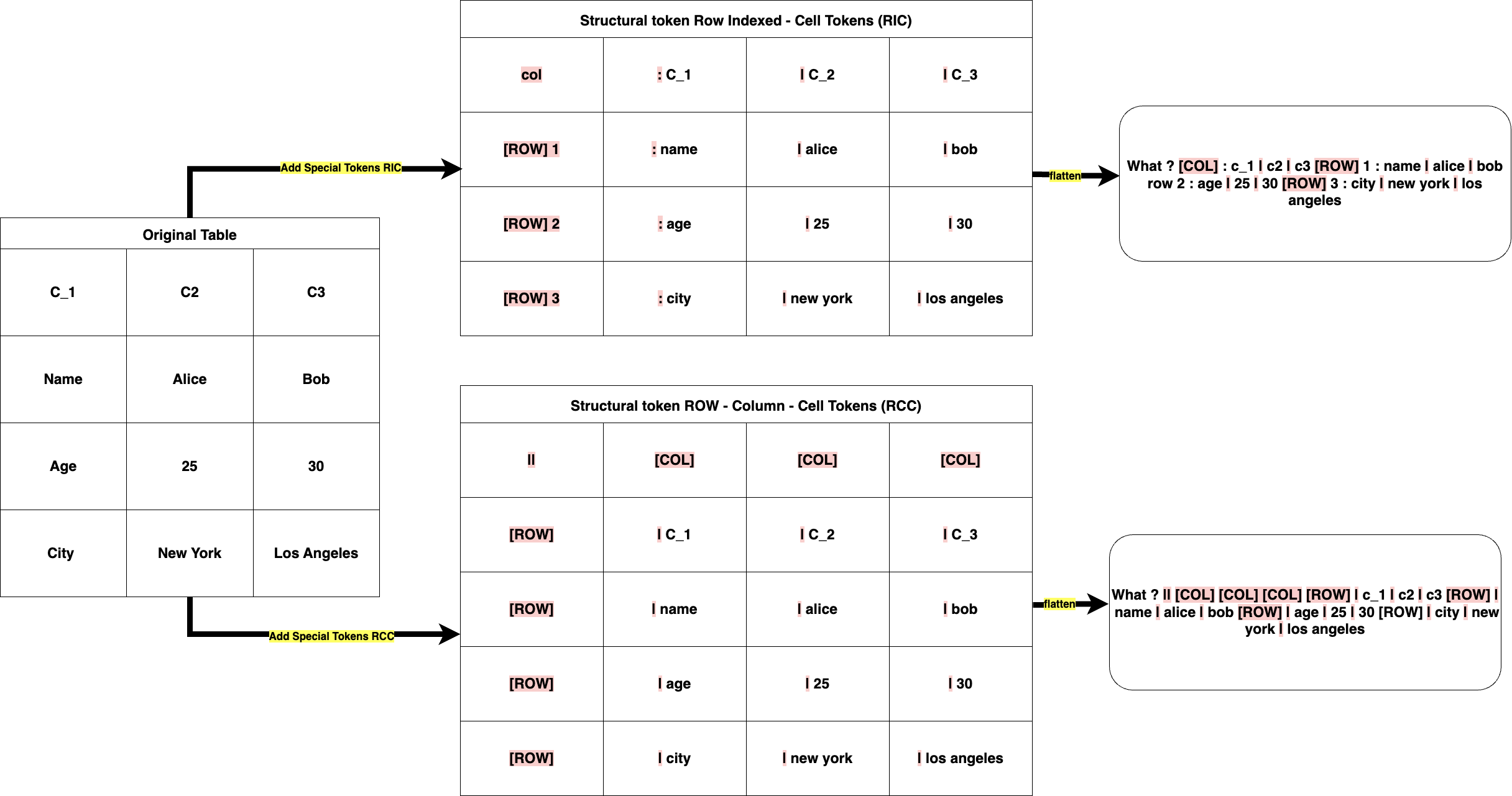}
  \caption{This figure illustrates two strategies for incorporating special tokens: T2 (Row-Column-Cell Tokens) and T1 (Row Indexed-Cell Tokens). In both approaches, special tokens are first added to the table, and then the table is flattened by concatenating each row sequentially.}
  \label{fig:specialtokens}
\end{figure*}

\end{document}